# Learning to Cooperate via Policy Search

**Leonid Peshkin**
pesha@ai.mit.edu
MIT AI Laboratory
545 Technology Square
Cambridge, MA 02139

**Kee-Eung Kim**
kek@cs.brown.edu
Computer Science Dept.
Brown University, Box 1910
Providence, RI 02906

**Nicolas Meuleau**
nm@ai.mit.edu
MIT AI Laboratory
545 Technology Square
Cambridge, MA 02139

**Leslie Pack Kaelbling**
lpk@ai.mit.edu
MIT AI Laboratory
545 Technology Square
Cambridge, MA 02139

## Abstract

Cooperative games are those in which both agents share the same payoff structure. Value-based reinforcement-learning algorithms, such as variants of Q-learning, have been applied to learning cooperative games, but they only apply when the game state is completely observable to both agents. Policy search methods are a reasonable alternative to value-based methods for partially observable environments. In this paper, we provide a gradient-based distributed policy-search method for cooperative games and compare the notion of local optimum to that of Nash equilibrium. We demonstrate the effectiveness of this method experimentally in a small, partially observable simulated soccer domain.

## 1  INTRODUCTION

The interaction of decision makers who share an environment is traditionally studied in game theory and economics. The game theoretic formalism is very general, and analyzes the problem in terms of solution concepts such as Nash equilibrium [12], but usually works under the assumption that the environment is perfectly known to the agents.

In reinforcement learning [8, 18], no explicit model of the environment is assumed, and learning happens through trial and error. Recently, there has been interest in applying reinforcement learning algorithms to multi-agent environments. For example, Littman [9] describes and analyzes a *Q-learning*-like algorithm for finding optimal policies in the framework of zero-sum Markov games, in which two players have strictly opposite interests. Hu and Wellman [7] propose a different multi-agent Q-learning algorithm for *general-sum* games, and argue that it converges to a Nash equilibrium.

A simpler, but still interesting case, is when multiple agents share the same objectives. A study of the behavior of agents employing Q-learning individually was made by Claus and Boutilier [6], focusing on the influence of game structure and exploration strategies on convergence to Nash equilibria. In Boutilier's later work [5], an extension of value iteration was developed that allows each agent to reason explicitly about the state of coordination.

However, all of this research assumes that the agents have the ability to completely and reliably observe both the state of the environment and the reward received by the whole system. Schneider et. al. [15] investigate a case of distributed reinforcement learning, in which agents have complete and reliable state observation, but only receive a local reinforcement signal. They investigate rules that allow individual agents to share reinforcement with their neighbors. In this paper we investigate the complementary problem in which the agents all receive the shared reward signal, but have incomplete, unreliable, and generally different perceptions of the world state. In such environments, value-search methods are generally inappropriate, causing us to turn to policy-search methods [19, 2, 3] which we have applied previously to single-agent partially observable domains [10, 13].

In this paper we describe a gradient-descent policy-search algorithm for cooperative multi-agent domains. In this setting, after each agent performs its action given its observation according to some individual strategy, they all receive the same payoff. Our objective is to find a learning algorithm that makes each agent independently find a strategy that enables the group of agents to receive the optimal payoff. Although this will not be possible in general, we present a distributed algorithm that finds *local* optima in the space of the agents' policies.

The rest of the paper is organized as follows. In section 2, we give a formal definition of a cooperative multi-agent environment. In section 3, we review the gradient descent algorithm for policy search, then develop it for the multi-agent setting. In section 4, we discuss the different notions of optimality for strategies. Finally, we present empirical results in section 5.

## 2 IDENTICAL PAYOFF GAMES

An *identical payoff stochastic game* (IPSG)[1] [11] describes the interaction of a set of agents with a Markov environment in which they all receive the same payoffs An (IPSG) is a tuple $\langle S, \pi_0^S, G, T, r \rangle$, where $S$ is a discrete state space; $\pi_0^S$ is a probability distribution over the initial state; $G$ is a collection of agents, where an *agent* $i$ is a 3-tuple, $\langle A^i, O^i, B^i \rangle$, of its discrete action space $A^i$, discrete observation space $O^i$, and observation function $B^i: S \to \mathcal{P}(O^i)$; $T: S \times \mathcal{A} \to \mathcal{P}(S)$ is a mapping from states of the environment and actions of the agents to probability distributions over states of the environment[2]; and $r: S \times \mathcal{A} \to \mathcal{R}$ is the payoff function, where $\mathcal{A} = \prod_i A^i$ is the joint action space of the agents. When all agents in $G$ have the identity observation function $B(s) = s$ for all $s \in S$, the game is *completely observable*. Otherwise, it is a *partially observable IPSG* (POIPSG).

In a POIPSG, at each time step: each agent $i \in (1..m)$ observes $o_i(t)$ corresponding to $B^i(s(t))$ and selects an action $a_k(t)$ according to its strategy; a compound action $\vec{a}(t) = (a_1(t), \ldots, a_m(t))$ from the joint action space $\mathcal{A}$ is performed, inducing a state transition of the environment; and the identical reward $r(t)$ is received by all agents.

The objective of each agent is to choose a strategy that maximizes the *value of the game*. For a discount factor $\gamma \in [0, 1)$ and a set of strategies $\vec{\mu} = (\mu_1, \ldots, \mu_m)$, given the distribution over initial state $\pi_0^S$, the value of the game is

$$V(\vec{\mu}, \pi_0^S) = \sum_{t=0}^{\infty} \gamma^t E(r(t) \mid \vec{\mu}, \pi_0^S). \qquad (1)$$

In the general case, a *strategy* for some agent is a mapping from the history of all observations from the beginning of the game into the current action $a(t)$. We limit our consideration in this paper to cases in which the agent's actions may depend only on the current observation, or in which the agent has a finite internal memory. When actions depend only on the current observation, the policy is called a *memoryless* or *reactive policy*. When this dependence is probabilistic, we call it a *stochastic reactive policy*, otherwise a *deterministic reactive policy*.

Note that in a completely observable IPSG, reactive policies are sufficient to implement the best possible joint strategy. This follows directly from the fact that every MDP has an optimal deterministic reactive policy [14]. Therefore an MDP with the product action space $\prod_i A^i$ corresponding to a completely observable IPSG also has one, representable by deterministic reactive policies for each agent. However, it has been shown that in partially observable environments, the best reactive policy can be arbitrarily worse than the best policy using memory [16]. This statement can also be easily extended to POIPSGs.

---
[1] IPSG's are also called *stochastic games* [7], Markov games [9] and *multi-agent Markov decision processes* [5].

[2] $\mathcal{P}(\Omega)$ denotes the set of probability distributions defined on some space $\Omega$.

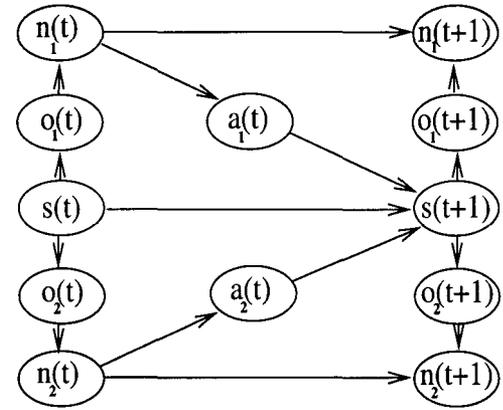

Figure 1: An influence diagram for two agents with FSCs in a POIPSG.

There are many possibilities for constructing policies with memory [13, 10]. In this work we use a *finite state controller* (FSC) for each agent. A more detailed description of FSCs and derivation of algorithms for learning them may be found in a previous paper [10]; we simply state the definition here.

A *finite state controller* (FSC) for an agent with action space $A$ and observation space $O$ is a tuple $\langle N, \pi_0^N, \eta, \psi \rangle$, where $N$ is a finite set of internal controller states, $\pi_0^N$ is a probability distribution over the initial internal state, $\eta : N \times O \to \mathcal{P}(N)$ is the internal state transition function that maps an internal state and observation into a probability distribution over internal states, and $\psi : N \to \mathcal{P}(A)$ is the action function that maps an internal state into a probability distribution over actions. Figure 1 depicts an influence diagram for two agents controlled by FSCs.

Note that in partially observable environments, agents controlled by FSCs might not have enough memory to even represent an optimal policy which could, in general, require infinite memory, as in a partially observable Markov decision process (POMDP) [17]. In this paper, we concentrate on the problem of finding the (locally) optimal controller from the class of FSCs with some fixed size of memory.

To better understand IPSGs, let us consider an example from Boutilier [5], illustrated in figure 2. There are two agents, $a_1$ and $a_2$, each of which has a choice of two actions, $a$ and $b$, at any of three states. All transitions are deterministic and are labeled by the joint action that corresponds to the transition. For instance, the joint action $(a, b)$ corresponds to the first agent performing action $a$ and the second agent performing action $b$. Here, $*$ refers to any action taken by the corresponding agent.

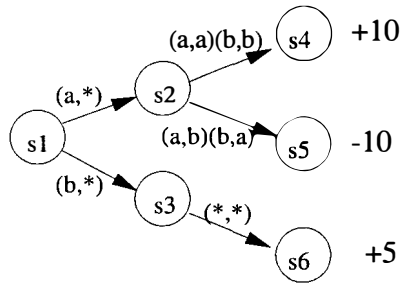

Figure 2: A coordination problem in a completely observable identical payoff game.

The starting state is $s1$, where the first agent alone decides whether to move the environment to state $s2$ by performing action $a$ or to state $s3$ by performing action $b$. In state $s3$, no matter what both agents do as the next step, they receive a reward of $+5$ in state $s6$ risk-free. In state $s2$, the agents have a choice of cooperating—choosing the same action, whether $(a,a)$ or $(b,b)$—with reward $+10$ in state $s4$, or not—choosing different actions, whether $(a,b)$ or $(b,a)$—and getting $-10$ in state $s5$.

We will represent a joint policy with parameters $p_{State}^{Agent}$, denoting the probability that an agent will perform action $a$ in the corresponding state. Only three parameters are important for the outcome: $\{p_1^1, p_2^1; p_2^2\}$. The optimal joint policies are $\{1,1;1\}$ or $\{1,0;0\}$, which are deterministic reactive policies.

## 3 GRADIENT DESCENT FOR POLICY SEARCH

In this section, we first introduce a general method for using gradient descent in policy spaces, then show how it can be applied to multi-agent problems.

### 3.1 Basic Algorithm

Williams introduced the notion of policy search for reinforcement learning in his REINFORCE algorithm [19, 20], which was generalized to a broader class of error criteria by Baird and Moore [2, 3].

We will start by considering the case of a single agent interacting with a POMDP. The agent's policy $\mu$ is assumed to depend on some internal state taking on values in finite set $N$. We will not make any further commitment to details of the policy's architecture, as long as it defines the probability of action given past history as a continuous differentiable function of some set of parameters $w$.

First, we will establish some notation. We denote by $H_t$ the set of all possible experience sequences of length $t$: $h = \langle n(0), o(1), n(1), a(1), r(1), \ldots, o(t), n(t), a(t), r(t), o(t+1)\rangle$. In order to specify that some element is a part of the history $h$ at time $\tau$, we write, for example, $r(\tau,h)$ and $a(\tau,h)$ for the $\tau^{th}$ reward and action in the history $h$. We will also use $h^\tau$ to denote a prefix of the sequence $h \in H_t$ truncated at time $\tau \leq t$: $h^\tau \stackrel{\text{def}}{=} \langle n(0), o(1), n(1), a(1), r(1), \ldots, o(\tau), n(\tau), a(\tau), r(\tau), o(\tau+1)\rangle$. The value defined by equation 1 can be rewritten as

$$V(\mu, \pi_0^S) = \sum_{t=1}^{\infty} \gamma^t \sum_{h \in H_t} \Pr(h \mid \pi_0^S, \mu) r(t, h). \quad (2)$$

Let us assume the policy is expressed parametrically in terms of a vector of weights $\vec{w} = \{w_1, \ldots, w_M\}$. If we could calculate the derivative of $V$ for each $w_k$, it would be possible to do an exact gradient descent on value $V$ by making updates $\Delta w_k = \alpha \frac{\partial}{\partial w_k} V$. We can compute the derivative for each weight $w_k$,

$$\frac{\partial V(\mu, \pi_0^S)}{\partial w_k} = \sum_{t=1}^{\infty} \gamma^t \sum_{h \in H_t} \left[ r(t,h) \frac{\partial \Pr(h \mid \pi_0^S, \mu)}{\partial w_k} \right]$$

$$= \sum_{t=1}^{\infty} \gamma^t \sum_{h \in H_t} \Pr(h \mid \pi_0^S, \mu) r(t,h)$$

$$\times \sum_{\tau=1}^{t-1} \frac{\partial}{\partial w_k} \ln\left(\Pr\left(a(\tau,h) \mid \pi_0^S, h^\tau, \mu\right)\right).$$

But, in the spirit of reinforcement learning, we cannot assume the knowledge of a world model that would allow us to calculate $\Pr(h \mid \pi_0^S, \mu)$, so we must retreat to stochastic gradient descent instead. We sample from the distribution of histories by interacting with the environment, and calculate during each trial an estimate of the gradient, accumulating the quantities:

$$\gamma^t r(t,h) \sum_{\tau=1}^{t-1} \frac{\partial \ln\left(\Pr\left(a(\tau,h) \mid \pi_0^S, h^\tau, \mu\right)\right)}{\partial w_k}, \quad (3)$$

for all $t$. For a particular policy architecture, this can be readily translated into a gradient descent algorithm that is guaranteed to converge to a local optimum of $V$.

### 3.2 Central Control Of Factored Actions

Now let us consider the case in which the action is factored, meaning that each action $\vec{a}$ consists of several components $\vec{a} = (a_1, \ldots, a_m)$. We can consider two kinds of controllers: a *joint controller* is a policy mapping observations to the complete joint distribution $\pi(\vec{a})$; a *factored controller* is made up of independent sub-policies $\mu_{a_i} : O^i \to \mathcal{P}(a_i)$ (possibly with a dependence on individual internal state) for each action component.

Factored controllers can represent only a subset of the policies represented by joint controllers. Obviously, any product of policies for the factored controller $\prod_i \mu_{a_i}$ can be represented by a joint controller $\mu_{\vec{a}}$, for which $\Pr(\vec{a}) = \prod_{i=1}^{N} \Pr(a_i)$. However, there are some stochastic joint controllers that cannot be represented by any factored controller, because they require coordination of probabilistic choice across action components, which we illustrate by the following example.

The first action component controls the liquid component of a meal $a_1 \in \{\text{vodka}, \text{milk}\}$ and the second controls the solid one $a_2 \in \{\text{pickles}, \text{cereal}\}$. For the sake of argument, let us assume that sticking to one combination or another is not as good as a "mixed strategy", meaning that for a healthy diet, we sometimes want to eat milk with cereal, other times vodka with pickles. The optimal policy is randomized, say 10% of the time $\vec{a} = (vodka, pickles)$ and 90% of the time $\vec{a} = (milk, cereal)$. But when the components are controlled independently, we cannot represent this policy. With randomization, we are forced to drink vodka with cereal or milk with pickles on some occasions.

Because we are interested in individual agents learning independent policies, we concentrate on learning the best factored controller for a domain, even if it is suboptimal in a global sense. Requiring a controller to be factored simply puts constraints on the class of policies, and therefore distributions $P(a | \mu, h)$, that can be represented. The stochastic gradient-descent techniques of the previous section can still be applied directly in this case to find local optima in the controller space. We will call this method *joint gradient descent*.

### 3.3 Distributed Control Of Factored Actions

The next step is to learn to choose action components not centrally, but under the distributed control of multiple agents. One obvious strategy would be to have each agent perform the same gradient-descent algorithm in parallel to adapt the parameters of its own local policy $\mu_{a_i}$. Perhaps surprisingly, this *distributed gradient descent* (DGD) method is very effective.

**Theorem 1** *For factored controllers, distributed gradient descent is equivalent to joint gradient descent.*

*Proof:* We will show that for both controllers the algorithm will be stepwise the same, so starting from the same point in the search space, on the same data sequence, the algorithms will converge to the same locally optimal parameter setting.

For a factored controller, $\vec{h}$ can be described as $\langle n_1(0), ..., n_m(0), o_1(1), ..., o_m(1), n_1(1), ..., n_m(1), a_1(1), ..., a_m(1), r(1), ...\rangle$. The corresponding history for an individual agent $i$ is $h_i = \langle n_i(0), o_i(1), n_i(1), a_i(1), r(1), ...\rangle$. It is clear that a collection $h_1...h_m$ of individual histories for each agent specifies the joint history $\vec{h}$.

The joint gradient descent algorithm requires that we draw sample histories from $\Pr(\vec{h} | \pi_0^S, \vec{\mu})$ and that we do gradient descent on $\vec{w}$ with a sample of the gradient at each time step $t$ in the history equal to

$$\gamma^t r(t, h) \sum_{\tau=1}^{t-1} \frac{\partial \ln \Pr(a(\tau, h) | \pi_0^S, \vec{h}^\tau, \mu)}{\partial w}.$$

Whether a factored controller is being executed by a single agent, or it is implemented by agents individually executing policies $\mu_{a_i}$ in parallel, joint histories are generated from the same distribution $\Pr(\vec{h} | \pi_0^S, \langle \mu_{a_1}, ..., \mu_{a_m}\rangle)$. So the distributed algorithm is sampling from the correct distribution.

Now, we must show that the weight updates are the same in the distributed algorithm as in the joint one. Let $\vec{w}_k = (w_k^0, ..., w_k^{M_k})$ be the set of parameters controlling action component $a_k$. Then

$$\frac{\partial}{\partial w_k^j} \ln(\Pr(a_l(\tau) | \pi_0^S, \vec{h}_l^\tau, \mu_l)) = 0 \text{ for all } k \neq l;$$

that is, the action probabilities of agent $l$ are independent of the parameters in other agents' policies. With this in mind, for factored controllers, the derivative in expression 3 becomes

$$\frac{\partial}{\partial w_k^j} \ln(\Pr(\vec{a}(\tau, h)) | \pi_0^S, \vec{h}^\tau, \vec{\mu})$$

$$= \frac{\partial}{\partial w_k^j} \ln \prod_{i=1}^{m} (\Pr(a_i(\tau, h)) | \pi_0^S, h_i^\tau, \mu_i)$$

$$= \sum_{i=1}^{m} \frac{\partial}{\partial w_k^j} \ln(\Pr(a_i(\tau, h)) | \pi_0^S, h_i^\tau, \mu_i)$$

$$= \frac{\partial}{\partial w_k^j} \ln(\Pr(a_k(\tau, h)) | \pi_0^S, h_k^\tau, \mu_k).$$

Therefore, the same weight updates will be performed by DGD as by joint gradient descent on a factored controller. ∎

This theorem shows that policy learning and control over component actions can be distributed among independent agents who are not aware of each others' choice of actions. An important requirement, though, is that agents perform simultaneous learning (which might be naturally synchronized by the coming of the rewards).

## 4 RELATING LOCAL OPTIMA TO NASH EQUILIBRIA

In game theory, the Nash equilibrium is a common solution concept. Because gradient descent methods can often be guaranteed to converge to local optima in the policy space, it is useful to understand how those points are related to Nash equilibria. We will limit our discussion to the two-agent case, but the results are generalizable to more agents.

A Nash equilibrium is a pair of strategies such that deviation by one agent from its strategy, assuming the other agent's strategy is fixed, cannot improve the overall performance. Formally, in an IPSG, a *Nash equilibrium* point is a pair of strategies $(\mu_1^*, \mu_2^*)$ such that:

$$V(\langle\mu_1^*, \mu_2^*\rangle, \pi_0^S) \geq V(\langle\mu_1, \mu_2^*\rangle, \pi_0^S),$$
$$V(\langle\mu_1^*, \mu_2^*\rangle, \pi_0^S) \geq V(\langle\mu_1^*, \mu_2\rangle, \pi_0^S)$$

for all $\langle\mu_1, \mu_2\rangle$. When the inequalities are strict, it is called a *strict Nash equilibrium*.

Every discounted stochastic game has at least one Nash equilibrium point [7]. It has been shown that under certain convexity assumptions about the shape of payoff functions, the gradient-descent process converges to an equilibrium point [1]. It is clear that the optimal Nash equilibrium point (the Nash equilibrium with the highest value) in an IPSG also is a possible point of convergence for the gradient descent algorithm, since it is the global optimum in the policy space.

Let us return to the game described in Figure 2. It has two optimal strict Nash equilibria at $\{1,1;1\}$ and $\{1,0;0\}$. It also has a set of sub-optimal Nash equilibria $\{0, p_2^1; p_2^2\}$, where $p_2^2$ can take on any value in the interval $[.25, .75]$ and $p_2^1$ can take any value in the interval $[0, 1]$. The sub-optimal Nash equilibria represent situations in which the first agent always chooses the bottom branch and the second agent acts moderately randomly in state $s2$. In such cases, it is strictly better for the first agent to stay on the bottom branch with expected value $+5$. For the second agent, the payoff is $+5$ no matter how it behaves, so it has no incentive to commit to a particular action in state $s2$ (which is necessary for the upper branch to be preferred).

In this problem, the Nash equilibria are also all local optima for the gradient descent algorithm. Unfortunately, this equivalence only holds in one direction in the general case. We state this more precisely in the following theorems.

**Theorem 2** *Every strict Nash equilibrium is a local optimum for gradient descent in the space of parameters of a factored controller.*

*Proof:* Assume that we have two agents and denote the strategy at the point of strict Nash equilibrium as $(\mu_1^*, \mu_2^*)$ encoded by parameter vector $\langle w_1^1 \ldots w_1^G, w_2^1 \ldots w_2^G\rangle$. For simplicity, let us further assume that $(\mu_1^*, \mu_2^*)$ is not on the boundary of the parameter space, and each weight is locally relevant: that is, that if the weight changes, the policy changes, too.

By the definition of Nash equilibrium, any change in value of the parameters of one agent without change in the other agent's parameters results in a decrease in the value $V$. In other words, we have that $\partial V / \partial w_i^j \leq 0$ and $-\partial V / \partial w_i^j \leq 0$ for all $j$ and $i$ at the equilibrium point. Thus, $\partial V / \partial w_i^j = 0$ for all $w_i^j$ at $(\mu_1^*, \mu_2^*)$, which implies it is a singular point of $V$. Furthermore, because the value decreases in every direction, it must be a maximum.

In the case of a locally irrelevant parameter $w_i^j$, $V$ will have a ridge along its direction. All points on the ridge are singular and, although they are not strict local optima, they are essentially local optima for gradient descent. ∎

The problem of Nash equilibria on the boundary of the parameter space is an interesting one. Whether or not they are convergence points depends on the details of the method used to keep gradient descent within the boundary. A particular problem comes up when the equilibrium point occurs when one or more parameters have infinite value (this is not uncommon, as we shall see in section 5). In such cases, the equilibrium cannot be reached, but it can usually be approached closely enough for practical purposes.

**Theorem 3** *Some local optima for gradient descent in the space of parameters of a factored controller are not Nash equilibria.*

*Proof:* Consider a situation in which each agent's policy has a single parameter $w_i$, so the policy space can be described by $\langle w_1, w_2 \rangle$. We can construct a value function $V(w_1, w_2)$ such that for some $c$, $V(\cdot, c)$ has two modes, one at $V(a, c)$ and the other at $V(b, c)$, such that $V(b, c) > V(a, c)$. Further assume that $V(a, \cdot)$ and $V(b, \cdot)$ each have global maxima $V(a, c)$ and $V(b, c)$. Then $V(a, c)$ is a local optimum that is not a Nash equilibrium. ∎

## 5 EXPERIMENTS

There are no established benchmark problems for multi-agent learning. To illustrate our method we present empirical results for two problems: the simple coordination problem of figure 2 and a small multi-agent soccer domain.

### 5.1 Simple Coordination Problem

We originally discussed policies for this problem using three weights, one corresponding to each of the $p_{State}^{Agent}$ probabilities. However, to force gradient descent to respect the bounds of the simplex, we used the standard Boltzmann encoding, so that for agent $i$ in state $s$ there are two weights $w_{s,a}^i$ and $w_{s,b}^i$, one for each action. The action probability is coded as a function of these weights as

$$p_s^i = \frac{e^{w_{sa}^i/\theta}}{e^{w_{sa}^i/\theta} + e^{w_{sb}^i/\theta}} > 0.$$

We ran DGD with a learning rate of $\alpha = .003$ and a discount factor of $\gamma = .99$; the results are shown in figure 3. The graph of a sample run illustrates how the agents typically initially move towards a sub-optimal policy. The policy in which the first agent always takes action $b$ and the second agent acts fairly randomly is a Nash equilibrium, as we saw in section 4. However, this policy is not exactly representable in the Boltzmann parameterization because it requires one of the weights to be infinite to drive a probability to either 0 or 1.

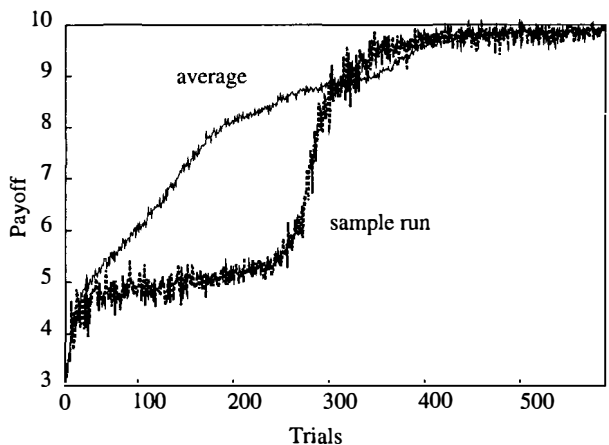

Figure 3: Average payoff and payoff of a sample run of a distributed gradient descent on the simple problem.

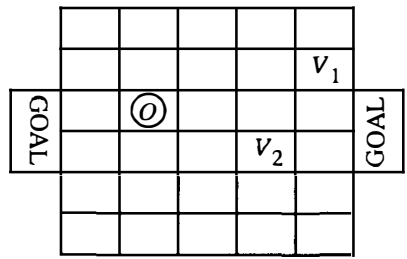

Figure 4: The soccer field. $V_1$ and $V_2$ represent learning agents and $O$ represents the opponent.

So, although the algorithm moves toward this policy, it never reaches it exactly. This means that there is an advantage for the second agent to drive its parameter toward 0 or 1, resulting in eventual convergence toward a global optimum (note that, in this parameterization, these optima cannot be reached exactly, either). The average of 10 runs shows that the algorithm always converges to a pair of policies with value very close to the maximum value of 10.

### 5.2 Soccer

We have also conducted experiments on a small soccer domain adapted from Littman [9]. The game is played on a $6 \times 5$ grid as shown in Figure 4. There are two learning agents on one team and a single opponent with a fixed strategy on the other. Every time the game begins, the learning agents are randomly placed in the right half of the field, and the opponent in the left half of the field. Each cell in the grid contains at most one player. Every player on the field (including the opponent) has an equal chance of initially possessing the ball.

At each time step, a player can execute one of the six actions: $\{North, South, East, West, Stay, Pass\}$. When an agent passes, the ball is transferred to the other agent on its team an the next time step. Once all players have selected actions, they are executed in a random order. When a player executes an action that would move it into the cell occupied by some other player, possession of the ball goes to the stationary player and the move does not occur. When the ball falls into one of the goals, the game ends and a reward of $\pm 1$ is given.

We made a partially observable version of the domain to test the effectiveness of DGD: each agent can only obtain information about which player possesses the ball and the status of cells to the north, south, east and west of its location. There are 3 possible observations for each cell: whether it is open, out of the field, or occupied by someone. In Figure 5, we compare the learning curves of DGD to those of Q-learning with a central controller for both the completely observable and the partially observable cases. We also show learning curves of DGD without the action *Pass* in order to measure the *cooperativeness* of the learned policies.

The graphs in the figure summarize simulations of the game against three different fixed-strategy opponents:
• Random: Executes actions uniformly at random.
• Greedy: Moves toward the player possessing the ball and stays there. Whenever it has the ball, it rushes to the goal.
• Defensive: Rushes to the front of its own goal, and stays or moves at random, but never leaves the goal area.

We show the average performance over 10 runs with error bars for standard deviation. The learning rate was 0.05 for DGD and 0.1 for Q-learning, and the discount factor was 0.999, throughout the experiments. Each agent in the DGD team learned a reactive policy. The policy's parameters were initialized by drawing uniformly at random from the appropriate domains. We used $\epsilon$-greedy exploration with $\epsilon = 0.4$ for Q-learning. The performance in the graph is reported by evaluating the greedy policy derived from the Q-table.

Because, in the completely observable case, this domain is an MDP (the opponent's strategy is fixed, so it is not really an adversarial game), Q-learning can be expected to learn the optimal joint policy, which it seems to do. It is interesting to note the slow convergence of completely observable Q-learning against the random opponent. We conjecture that this is because, against a random opponent, a much larger part of the state space is visited. The table-based value function offers no opportunity for generalization, so it requires a great deal of experience to converge.

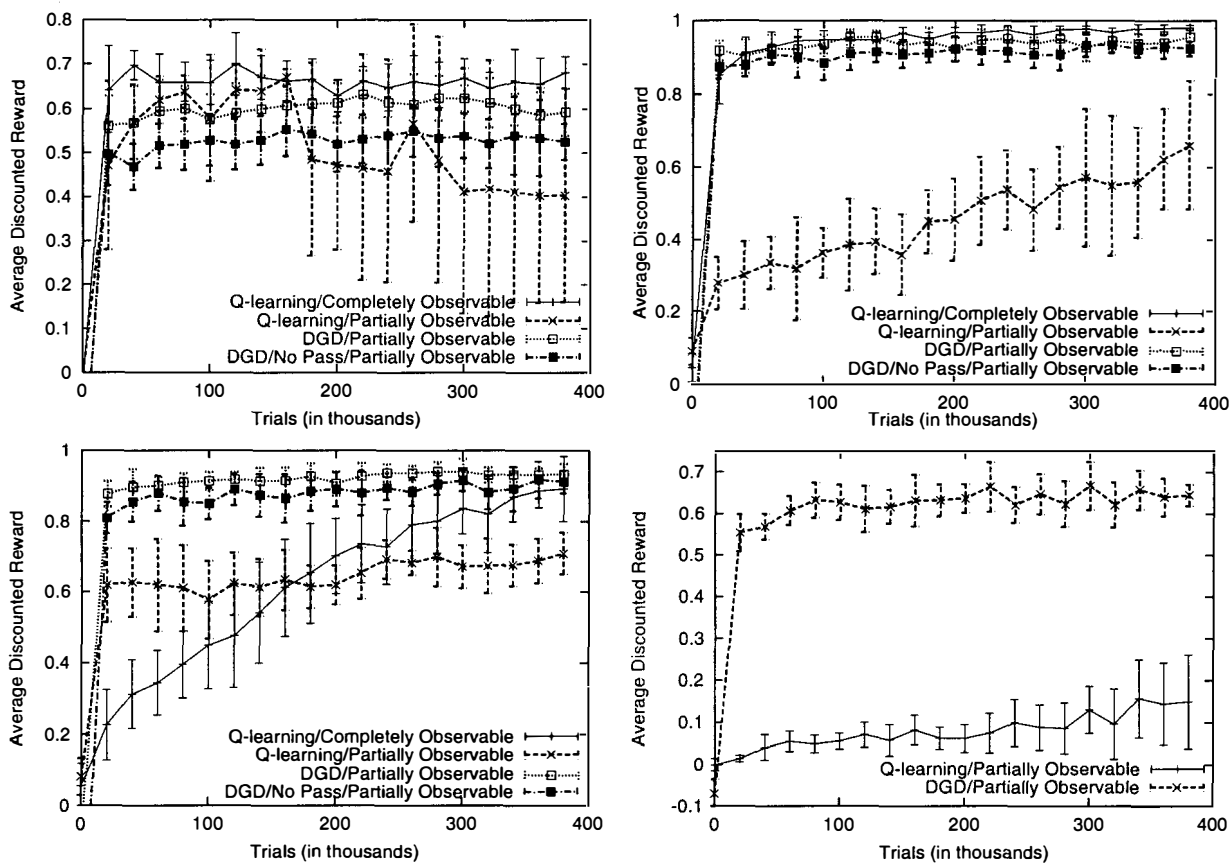

Figure 5: Learning curves of DGD (policy search) and Q-learning against defensive opponent (top left), greedy opponent (top right), random opponent (bottom left), and the team with two agents with different fixed strategies (bottom right).

As soon as observability is restricted, Q-learning no longer reliably converges to the best strategy. The joint Q-learner now has as its input the two local observations of the individual players. It behaves quite erratically, with extremely high variance because it sometimes converges to a good policy and sometimes to a bad one. This unreliable behavior can be attributed to the well-known problems of using value-function approaches, and especially Q-learning, in POMDPs.

The individual DGD agents have stochasticity in their action choices, which gives them some representational abilities unavailable to the Q-learner. We tried additional experiments in which the DGD agents had 4-state FSC's. Their performance did not improve appreciably. We expect that, in future experiments on a larger field with more players, it will be important for the agents to have internal state.

Despite the fact that they learn independently, the combination of policy search plus a different policy class allows them to gain considerably improved performance. We cannot tell how close this performance is to the optimal performance with partial observability, because it would be computationally impractical to solve the POIPSG exactly. Bernstein et. al. [4] show that in the finite-horizon case two-agent POIPSGs are *harder* to solve than POMDPs (in the worst-case complexity sense).

It is also important to see that the two DGD agents have learned to cooperate in some sense: when the same algorithm is run in a domain without the "pass" action, which allows one agent to give the ball to its teammate, performance deteriorates significantly against both defensive and greedy opponents. Since the agents don't know where they are with respect to the goal, they probably choose to pass whenever they are faced with the opponent. Against a completely random opponent, both strategies do equally well. It is probably sufficient, in this case, to simply run straight for the goal, so cooperation is not necessary.

We performed some additional experiments in a two-on-two domain in which one opponent behaved greedily and the other defensively. In this domain, the completely observable state space is so large that it is difficult to even store the Q table, let alone populate it with reasonable val-ues. Thus, we just compare two 4-state DGD agents with a limited-view centrally controlled Q-learning algorithm. Not surprisingly, we find that the DGD agents are considerably more successful.

Finally, we performed informal experiments with an increasing number of opponents. The opponent team was made up of one defensive agent and an increasing number of greedy agents. For all cases in which the opponent team had more than two greedy agents, DGD led to a defensive strategy in which, most of the time, the agents all rushed to the front of their goal and stayed there forever.

# 6 CONCLUSIONS AND FUTURE WORK

We have presented an algorithm for distributed learning in cooperative multi-agent domains. It is guaranteed to find local optima in the space of factored policies. We cannot show, however, that it always converges to a Nash equilibrium, because there are local optima in policy space that are not Nash equilibria. The algorithm performed well in a small simulated soccer domain.

It will be important to apply this algorithm in more complex domains, to see if the gradient remains strong enough to drive the search effectively, and to see whether local optima become problematic. An interesting extension of this work would be to allow the agents to perform explicit communication actions with one another to see if they are exploited to improve performance in the domain.

In addition, there may be more interesting connections to establish with game theory, especially in relation to solution concepts other than Nash equilibrium, which may be more appropriate in cooperative games.

**Acknowledgement**

Thanks to Craig Boutilier and Michael Schwarz for many relevant ideas and comments. Leonid Peshkin was supported by grants from NSF and NTT; Nicolas Meuleau in part by research grant from NTT; Kee-Eung Kim in part by AFOSR/RLF 30602-95-1-0020; and Leslie Pack Kaelbling in part by a grant from NTT and in part by DARPA Contract #DABT 63-99-1-0012.